\documentclass[12pt]{spieman} 
\usepackage{tabularx}
\usepackage{adjustbox}
\usepackage{amsmath,amsfonts,amssymb}
\usepackage{booktabs}
\usepackage{colortbl}
\usepackage{afterpage}
\definecolor{deepgreen}{RGB}{10,90,40} 
\definecolor{lessred}{RGB}{210,100,70}  
\definecolor{lessblue}{RGB}{100,100,230} 
\definecolor{lightgray}{gray}{0.8} 
\usepackage[colorlinks=true, allcolors=blue]{hyperref}
\usepackage{pifont}
\usepackage{graphicx}
\usepackage{setspace}
\usepackage{tocloft}
\usepackage{textcomp}
\usepackage{color,soul}
\usepackage{subcaption}
\usepackage{float}
\usepackage[colorlinks=true, allcolors=blue]{hyperref}
\usepackage{multirow}
\usepackage[table,xcdraw,dvipsnames]{xcolor}
\usepackage{lineno}
\usepackage{makecell}
\usepackage{textcomp}
\usepackage{array}
\usepackage{listings}
\usepackage{mathptmx}
\usepackage{lineno}
\usepackage{epsfig}
\usepackage{times}
\usepackage{rotating}
\usepackage{makeidx}
\usepackage{url}
\usepackage{booktabs}
\usepackage{tabularx}
\usepackage{blindtext}
\usepackage{listings}

\colorlet{Mycolor1}{green!10!orange}

\title{GloFinder: AI-empowered QuPath Plugin for WSI-level Glomerular Detection, Visualization, and Curation}
\author[a]{Jialin Yue}  
\author[b]{Tianyuan Yao}
\author[b]{Ruining Deng}
\author[a]{Siqi Lu}
\author[a]{Junlin Guo}
\author[b]{Quan Liu}
\author[c]{Mengmeng Yin}
\author[a]{Juming Xiong}
\author[c]{Haichun Yang}
\author[a,b,c,*]{Yuankai Huo}
\affil[a]{Department of Electrical and Computer Engineering, Vanderbilt University Medical Center, Nashville, TN, USA}
\affil[b]{Department of Computer Science, Vanderbilt University, Nashville, TN, USA}
\affil[c]{Department of Pathology, Microbiology and Immunology, Vanderbilt University Medical Center, Nashville, TN, USA}

\cftpagenumbersoff{figure}
\cftpagenumbersoff{table} 
\begin{document} 
\begin{sloppypar}
\maketitle

\begin{abstract}
Artificial intelligence (AI) has demonstrated significant success in automating the detection of glomeruli—key functional units of the kidney—from whole slide images (WSIs) in kidney pathology. However, existing open-source tools are often distributed as source code or Docker containers, requiring advanced programming skills that hinder accessibility for non-programmers, such as clinicians. Additionally, current models are typically trained on a single dataset and lack flexibility in adjusting confidence levels for predictions. To overcome these challenges, we introduce GloFinder, a QuPath plugin designed for single-click automated glomeruli detection across entire WSIs with online editing through the graphical user interface (GUI). GloFinder employs CircleNet, an anchor-free detection framework utilizing circle representations for precise object localization, with models trained on approximately 160,000 manually annotated glomeruli. To further enhance accuracy, the plugin incorporates Weighted Circle Fusion (WCF)—an ensemble method that combines confidence scores from multiple CircleNet models to produce refined predictions, achieving superior performance in glomerular detection. GloFinder enables direct visualization and editing of results in QuPath, facilitating seamless interaction for clinicians and providing a powerful tool for nephropathology research and clinical practice.

\end{abstract}


{\noindent \footnotesize\textbf{*Corresponding Author:} Yuankai Huo,  \linkable{yuankai.huo@vanderbilt.edu} }

\begin{spacing}{2}

\section{Introduction}

The digitization of histological slides has significantly advanced the field of computational pathology, enabling the application of sophisticated image analysis techniques to whole-slide images (WSIs). Accurate detection and segmentation of glomeruli—key structures in renal pathology—are essential for diagnosing and understanding kidney diseases. Automated glomeruli detection not only enhances the efficiency of pathological assessments but also improves the consistency and reproducibility of diagnoses, which are critical for effective patient care and research.

Many deep learning based automatic glomerular detection tools, such as CircleNet~\cite{nguyen2021circle}, have emerged as powerful tools for detecting circular glomeruli. For example, CircleNet leverages geometric properties to enhance detection accuracy by predicting the centers and radii of circles that best encapsulate these objects. However, existing open-source AI glomerular detection tools are distributed as source code or Docker containers, making them inaccessible to clinicians and other users without advanced programming expertise. This limitation hampers its seamless integration into routine pathological workflows, thereby restricting its potential to assist doctors effectively in disease analysis.

\begin{figure}[ht]
    \centering
    \includegraphics[width=\linewidth]{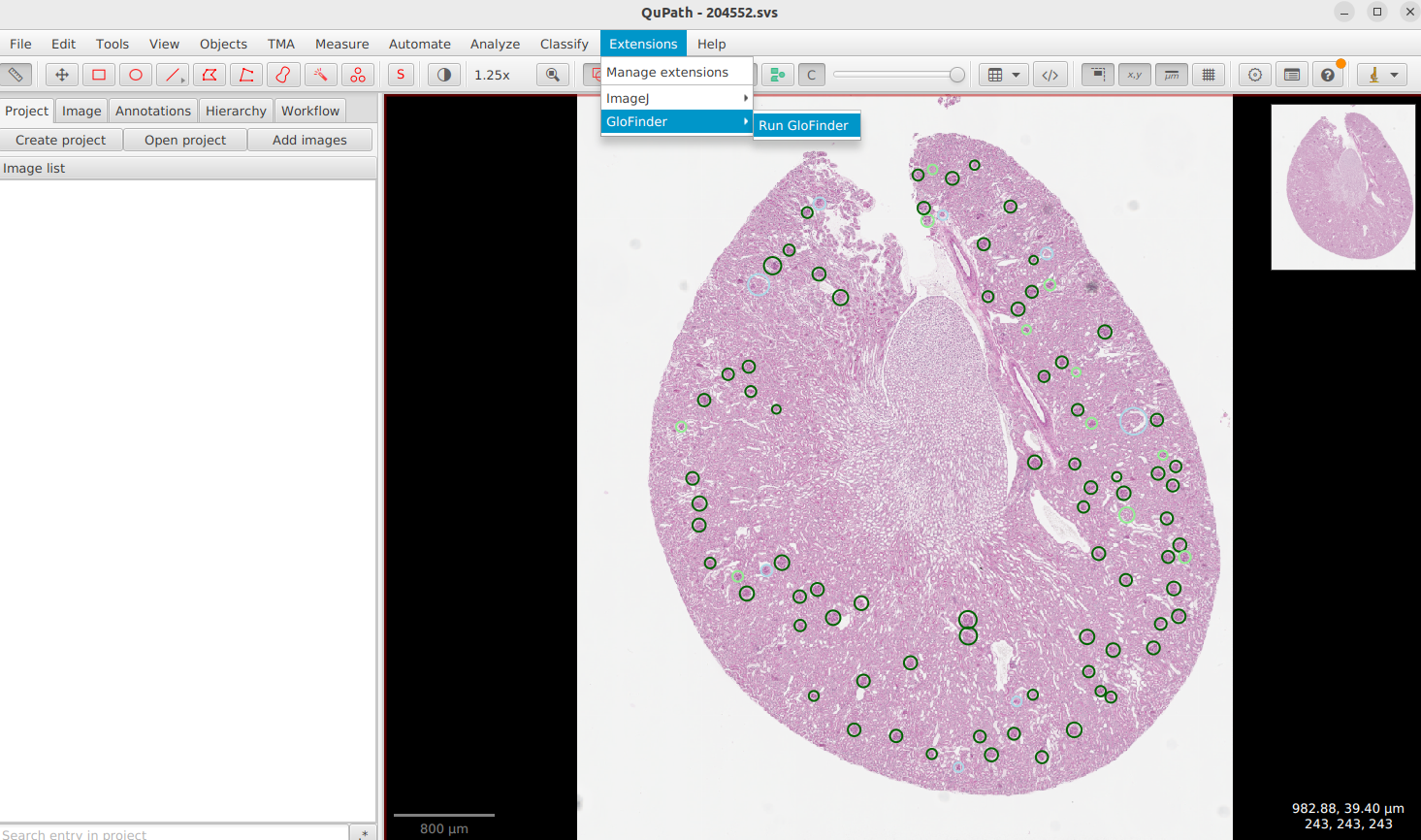}
    \caption{\textbf{Glomerular detection results using the GloFinder plugin.} Detected glomeruli are represented as circles with varying colors indicating detection confidence.}
    \label{fig1}
\end{figure}

To address these challenges, we present GloFinder, a AI-empowered QuPath~\cite{bankhead2017qupath} plugin, which is designed to facilitate the fully automatic glomerular detection (1) at WSI-level, (2) with a single click in graphical user interface (GUI), and (3) with online editing. This plugin streamlines the detection of glomeruli in WSIs, allowing clinicians to perform automated detection without any programming skills. Central to our approach is to train the state-of-the-art (SOTA) CircleNet method with 160,000 manually annotated glomeruli from both in-house and public datasets~\cite{yue2024weighted}, complemented by the introduction of Weighted Circle Fusion (WCF)~\cite{yue2024weighted}. WCF aggregates predictions from multiple CircleNet models by weighting and merging overlapping circles based on their confidence scores, thereby improving the robustness and reliability of the final annotations. The plugin offers a user-friendly GUI interface within the QuPath environment, enabling clinicians to detect glomeruli across an entire WSI with a single click. An illustration of the final detection results displayed within QuPath is shown in Figure~\ref{fig1}, where detected glomeruli are overlaid on the WSI. After applying the Weighted Circle Fusion, our plugin visualizes glomeruli detected by different numbers of models using distinct colors, which will be automatically categorized within the QuPath annotation menu. This feature allows clinicians to focus on glomeruli with fewer fused detections, which are more likely to be incorrect, facilitating targeted verification and correction. Additionally, the workflow is designed to be intuitive, facilitating easy visualization, annotation, and editing of detected glomeruli. Beyond glomeruli detection, the plugin's architecture is adaptable for identifying other circular biomedical objects, such as cell nuclei, making it a versatile pipeline for other potential applications in medical image analysis.

Our contributions in this paper are threefold:

$\bullet$ We develop and disseminate GloFinder, an open-source AI-powered QuPath~\cite{bankhead2017qupath} plugin, which is designed to enable fully automated glomerular detection with the following key features: (1) whole slide image (WSI)-level analysis, (2) single-click operation via a graphical user interface (GUI), and (3) seamless online editing capabilities.

$\bullet$ GloFinder achieves over a 5\% improvement in detection performance by employing WCF on multiple CircleNet models with over 160,000 manually annotated training samples.

$\bullet$ Using GloFinder within a human-in-the-loop annotation strategy reduces annotation time by 68.59\% compared to the current data curation pipeline.

\section{Methods}

This section outlines the methodologies employed in developing the GloFinder, a QuPath plugin designed for automated glomeruli detection. The methodologies are divided into two primary segments:the GloFinder plugin capacities, and the implementation of CircleNet and Weighted Circle Fusion for glomeruli detection.

\subsection{Plugin Capacities}

The GloFinder plugin prioritizes simplicity and ease of use, seamlessly integrating into the QuPath environment to facilitate automated glomeruli detection in whole-slide images (WSIs). Installation is straightforward: users simply open QuPath and drag the plugin's JAR file into the interface. Once installed, the plugin becomes accessible via the QuPath extensions menu, where users can initiate the detection process with a single click. The plugin supports multiple WSI formats, including commonly used types such as \texttt{.svs} and \texttt{.scn}, ensuring broad compatibility across different imaging datasets.

Beyond its user-friendly interface, GloFinder enhances flexibility and interactivity. After processing, the plugin saves the Python scripts used for detection, the trained models, and the generated GeoJSON files directly to the user's desktop. This approach allows users to access, modify, or adjust the detection code and results according to their specific needs, facilitating customization and further development. Additionally, the plugin visualizes glomeruli detected by varying numbers of models using distinct colors and categorizes them within the QuPath annotation menu. This feature enables clinicians to easily identify and focus on glomeruli with fewer supporting detections—which are more likely to be inaccurate—thereby aiding targeted verification and correction.

\begin{figure}[H]
    \centering
    \includegraphics[width=0.8\linewidth]{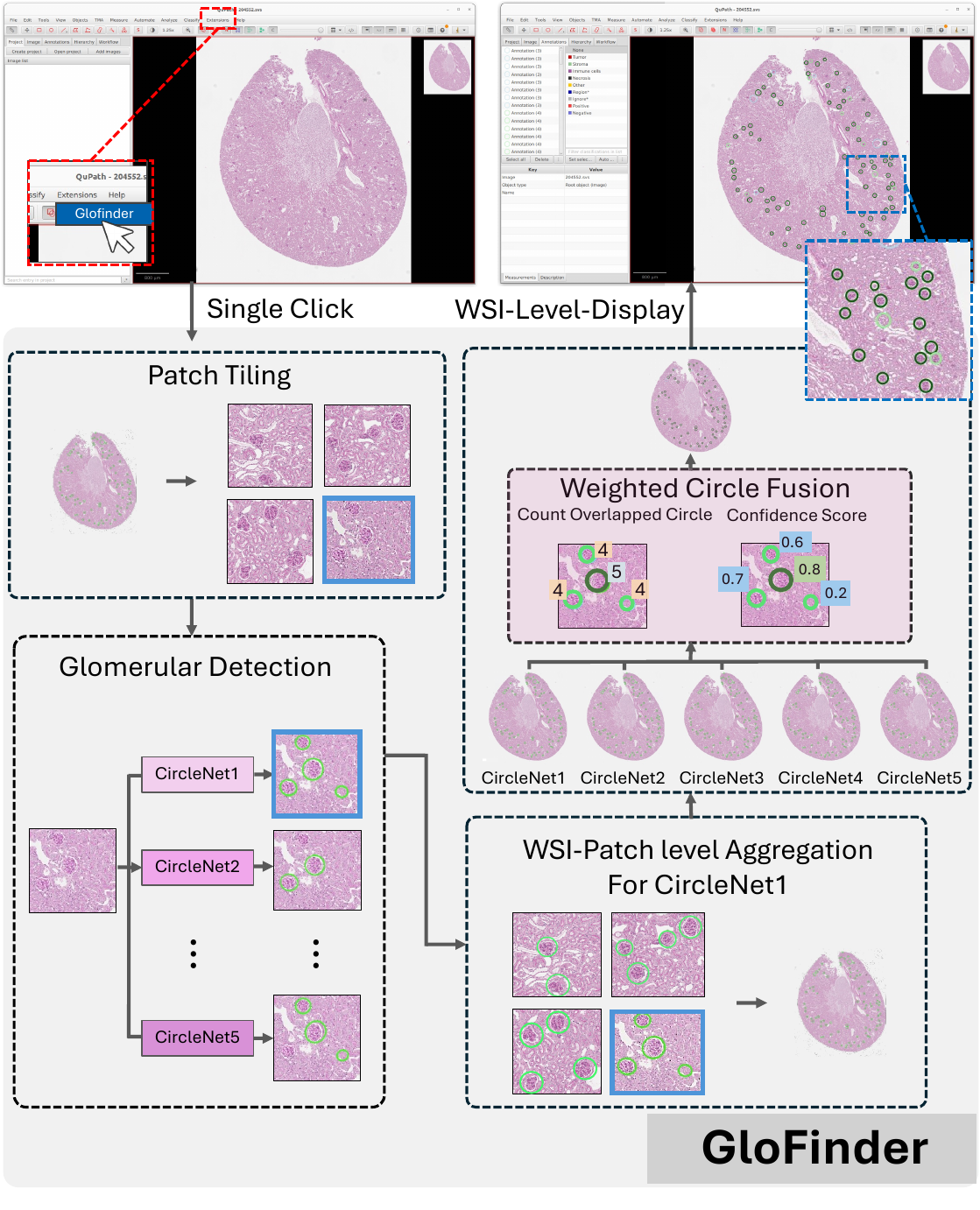}
    \caption{\textbf{The workflow of the GloFinder plugin's internal algorithm.} GloFinder first tiles the WSI into overlapping patches. Five CircleNet models, each trained on different datasets, detect glomeruli within these patches. The detection results are then aggregated back into the original WSI space. The Weighted Circle Fusion algorithm is applied to merge detections and enhance accuracy. Finally, the fused results are presented and displayed within the QuPath interface. This entire process only requires a single click on the GloFinder button from the extension menu.}
    \label{fig2}
\end{figure}

\subsection{CircleNet and Weighted Circle Fusion}

In our plugin, the detection algorithm begins by segmenting the WSI into multiple overlapping patches, each overlapping by half of its area with adjacent patches. This strategy increases the likelihood of capturing entire glomeruli that may reside at the edges of patches, thereby enhancing detection accuracy and reducing boundary artifacts. The overlapping patches ensure that glomeruli located near the boundaries are not missed during detection, as they are included in multiple patches.

Each patch is then processed by five CircleNet models, which are specialized detection networks designed to identify circular structures by predicting the center coordinates and radii of circles representing objects such as glomeruli. CircleNet adopts a rotation-consistent circle representation, making it particularly adept at detecting spherical structures in medical images. After the detection process, the relative coordinates of the detected glomeruli within each patch are transformed back to their corresponding positions on the entire WSI. This coordinate transformation is crucial for accurately mapping the detections onto the original image.

To eliminate redundant detections resulting from overlapping patches and multiple model predictions, the Non-Maximum Suppression (NMS) algorithm is applied. NMS filters out overlapping detections by retaining only the one with the highest confidence score for each glomerulus, ensuring that each object is represented by a single detection result. After obtaining five different sets of detection results from the five CircleNet models, we employ the Weighted Circle Fusion algorithm to fuse these results and improve accuracy. WCF is an ensemble method that combines the outputs of multiple models by weighting and merging overlapping circles based on their confidence scores, effectively leveraging the strengths of each model to enhance detection precision~\cite{yue2024weighted}.

By integrating these steps—patch-based processing, CircleNet detection, coordinate transformation, NMS filtering, and result fusion using WCF—we obtain the final detection results on the entire WSI. This comprehensive approach not only improves detection accuracy but also enhances the robustness of the plugin in various clinical scenarios. Figure~\ref{fig2} illustrates the workflow of our detection method, highlighting each stage of the process.

\section{Experiments}

\subsection{Training Data Configuration}

The models were developed and trained using an in-house dataset of murine glomeruli. Each CircleNet model was trained on a dataset of over 160,000 manually annotated glomeruli. To enhance learning diversity, approximately 30,000 glomeruli varied between the training datasets of different models, allowing the ensemble to better generalize across varied data.

The training patches, each containing at least one glomerulus, were standardized to dimensions of 512 × 512 pixels through cropping or resizing. To improve model robustness and prevent overfitting, extensive data augmentation techniques were applied, including random rotations, scaling, and brightness adjustments.

For testing and evaluation, we utilized an independent dataset consisting of 15 PAS-stained WSIs, containing a total of 2,051 mouse glomeruli. This diverse test set allowed for an independent assessment of the model's performance.

\subsection{Model Training and Parameters}

The models were built upon the CircleNet architecture, utilizing a DLA-34 backbone~\cite{nguyen2021circle} to ensure robust feature extraction and accurate detection. Training for each model was conducted over 30 epochs, with slightly varied datasets used for each model to introduce diversity, thereby enhancing the ensemble's overall performance and generalizability.

After initial detection by the CircleNet models, the outputs were refined using the NMS algorithm. This step was essential to eliminate redundant and overlapping detections within the output of each individual model, ensuring cleaner and more precise results.

To further improve detection accuracy, the refined outputs from the individual models were combined using the WCF ensemble method. This technique leveraged the complementary strengths of multiple models by averaging their outputs based on confidence scores. The WCF method applied carefully chosen thresholds: the ``T count" threshold was set to 2, ensuring that a detection needed consensus from at least two models, and the default ``T score" threshold was set to 0.9, requiring high confidence for the averaged predictions to be considered valid. This ensemble approach significantly improved detection reliability and reduced false positives, making the system more robust for clinical and research applications.

\subsection{Evaluation Metrics}

The models were evaluated based on the mean Average Precision (mAP)~\cite{zhu2004recall} at Intersection over Union (IoU) values of 0.5 and 0.75. Additionally, mAP was computed across IoU thresholds ranging from 0.5 to 0.95 in increments of 0.05. The average recall across these IoU thresholds was also measured.

Given that the predictions utilize circle representations rather than traditional bounding boxes, we employed the circle Intersection over Union (cIoU)~\cite{nguyen2021circle} metric for evaluation. This metric calculates the ratio of the overlap area to the combined area of the predicted and ground truth circles.

\subsection{Computational Environment and Runtime Performance}

All experiments were conducted on a workstation equipped with an 8-core Intel Xeon W-2245 processor and an NVIDIA RTX A5000 GPU, running Ubuntu 22.04. The combination of a powerful CPU and GPU facilitated efficient processing of the computational tasks associated with glomeruli detection and analysis. The GloFinder plugin required an average of 21 seconds to process a single WSI, demonstrating its capability to perform rapid analysis suitable for clinical workflows.

\section{Results}

\subsection{Performance on Automatic Glomerular Detection}

In our experiments, we compared the performance of WCF method with individual CircleNet model predictions and other ensemble methods, NMS and Soft-NMS, on glomerular detection. Detailed performance metrics are presented in Table~\ref{table:fusionresult}, highlighting the superior mAP values achieved by our WCF method compared to individual models and other ensemble techniques. The result shows that our WCF method achieved significantly higher mAP values across various IoU thresholds. This improvement demonstrates the effectiveness of incorporating varied training data to enhance model robustness and generalizability.

\begin{table}[h]
\centering
\resizebox{0.9\textwidth}{!}{
\begin{tabular}{lcccc}
\toprule
Model & mAP(0.5:0.95) & mAP(@0.5IOU) & mAP(@0.75IOU) & Average Recall(0.5:0.95) \\ \hline\hline
CircleNet \#1~\cite{nguyen2021circle}   & 0.741   & {0.893}   & 0.830  & 0.729   \\ 
CircleNet \#2   &0.730   & 0.876   & 0.813   & 0.772 \\ 
CircleNet \#3  & 0.724  & 0.932  & 0.833  & 0.652  \\ 
CircleNet \#4  & \textbf{\textcolor{lessblue}{0.789}}  & \textbf{\textcolor{lessblue}{0.953}}  & \textbf{\textcolor{lessblue}{0.862}}  & 0.699  \\ 
CircleNet \#5  & 0.731  & 0.841  & 0.787  & \textbf{\textcolor{lessred}{0.836}}  \\ 
model avg.   &\textcolor[gray]{0.3}{\textbf{ 0.743}} &\textcolor[gray]{0.3}{\textbf{ 0.899}} &\textcolor[gray]{0.3}{\textbf{ 0.825}} &\textcolor[gray]{0.3}{\textbf{ 0.738 }}\\
\hline
NMS~~\cite{1699659}  & 0.644  & 0.749  & 0.696  & \textbf{\textcolor{lessblue}{0.834}}  \\ 
Soft-NMS~~\cite{bodla2017soft}  & 0.419  & 0.513  & 0.452  & 0.793 \\ 
GloFinder  & \textbf{\textcolor{lessred}{0.829}}  & \textbf{\textcolor{lessred}{0.955}}  & \textbf{\textcolor{lessred}{0.905}}  & 0.782  \\ \bottomrule
\end{tabular}
}
\caption{The table provides a detailed comparison of performance metrics for individual models, the average results of five models (highlighted in bold), and the results using NMS, Soft-NMS, and WCF fusion methods. Metrics such as mAP at various IoU thresholds and average recall are included. The bold numbers represent the average performance of the five models, while the highest values for each evaluation metric are marked in red, and the second-highest values are marked in blue.}
\label{table:fusionresult}
\end{table}

\subsection{Efficiency of Human-in-the-loop Annotation}

To evaluate the efficiency of manual annotation compared to a HITL approach, we conducted a time analysis for annotating 10 WSIs. The results demonstrated that the HITL method considerably improves annotation efficiency, requiring an average of 2.9 minutes per image compared to 9.23 minutes per image for manual annotation.

This significant reduction in annotation time, approximately 68.6\%, highlights the practical benefits of integrating our plugin into clinical workflows. By automating the initial detection of glomeruli and allowing clinicians to focus on reviewing and correcting annotations rather than creating them from scratch, the HITL approach enhances efficiency without compromising accuracy.

\section{Discussion}

The GloFinder plugin advances automated glomeruli detection in WSIs, offering practical benefits for clinical and research settings. It efficiently processes entire WSIs at once, detecting all glomeruli and displaying results directly within the QuPath interface. This capability eliminates the need for manual region selection or patch-based analysis, saving time and reducing potential oversights.

Its user-friendly design allows clinicians and pathologists to operate the plugin with a single click, seamlessly integrating results into QuPath for easy visualization, review, and modification. This simplicity streamlines workflows and encourages greater adoption of automated analysis tools in routine pathological assessments.

GloFinder is also flexible and adaptable. By downloading the detection code, models, and annotations to the local system, users can modify or customize these components as needed. The plugin can detect other circular biomedical objects by simply replacing the model, expanding its applicability to various medical imaging tasks.

However, limitations include the detection time required to process full WSIs and reliance on local execution, which demands sufficient hardware and may limit accessibility for users with less powerful machines. Future work will focus on optimizing the algorithm to reduce detection time and exploring cloud-based execution to offload computations from local systems, alleviating hardware constraints and enabling centralized management of updates.

In conclusion, GloFinder provides clinicians and clinical scientists a user-friendly option for automated glomeruli detection, offering ease of use, flexibility, and adaptability. By addressing current limitations through algorithm optimization and potential cloud integration, the plugin can become an even more powerful tool for medical image analysis, ultimately contributing to improved diagnostic workflows and patient outcomes.

\section{Conclusions}
In this study, we introduced GloFinder, a new QuPath plugin that streamlines automated glomeruli detection in whole slide images. By integrating the advanced CircleNet framework and the WCF ensemble method, GloFinder achieves state-of-the-art detection accuracy while providing an intuitive and accessible user interface for clinicians and researchers. Beyond fully automatic detection, the plugin is able to reduce annotation time through a human-in-the-loop workflow and enables direct visualization, editing, and customization within the QuPath environment.



\section*{Disclosures}
The authors of the paper have no conflicts of interest to report. 

\section*{Code and Demo Video}
The code used to generate the plugin is available in a Github repository (https://github.com/hrlblab/PathVisual) The demo video is also available in the same Github repository.

\section* {Acknowledgments}
This research was supported by NIH R01DK135597 (Huo), DoD HT9425-23-1-0003 (HCY), NIH NIDDK DK56942 (ABF). This work was also supported by Vanderbilt Seed Success Grant, Vanderbilt Discovery Grant, and VISE Seed Grant. This project was supported by The Leona M. and Harry B. Helmsley Charitable Trust grant G-1903-03793 and G-2103-05128. This research was also supported by NIH grants R01EB033385, R01DK132338, REB017230, R01MH125931, and NSF 2040462. We extend gratitude to NVIDIA for their support by means of the NVIDIA hardware grant. This work was also supported by NSF NAIRR Pilot Award NAIRR240055.
 

\bibliography{report}   
\bibliographystyle{spiejour}   


\vspace{2ex}\noindent\textbf{First Author} is currently a master student in Electrical and Computer Engineering at Vanderbilt University. She is advised by Prof. Yuankai Huo at HRLB Lab. Her main research interests are medical image analysis, deep learning, and computer vision. 

\vspace{1ex}
\noindent Biographies and photographs of the other authors are not available.

\listoffigures
\listoftables

\end{spacing}
\end{sloppypar}
\end{document}